\documentclass[11pt,a4paper]{article}
\usepackage[hyperref]{naaclhlt2019}
\usepackage{times}
\usepackage{latexsym}
\usepackage{url}
\usepackage{amsmath}
\usepackage{amssymb}
\usepackage{booktabs}
\usepackage{graphicx}
\usepackage{subcaption}
\usepackage{framed}
\usepackage{float}
\usepackage{resizegather}
\usepackage{tikz}

\DeclareMathOperator*{\argmax}{arg\,max}

\aclfinalcopy 
\def\aclpaperid{2035} 

\newcommand\blfootnote[1]{%
  \begingroup
  \renewcommand\thefootnote{}\footnote{#1}%
  \addtocounter{footnote}{-1}%
  \endgroup
}

\title{What's in a Name? \\ Reducing Bias in Bios without Access to Protected Attributes}

\author{Alexey Romanov$^1$, Maria De-Arteaga$^2$, Hanna Wallach$^3$, \\ \textbf{Jennifer Chayes$^3$, Christian Borgs$^3$, Alexandra Chouldechova$^2$,} \\ \textbf{Sahin Geyik$^4$, Krishnaram Kenthapadi$^4$, Anna Rumshisky$^1$, Adam Tauman Kalai$^3$} \\
$^1$University of Massachusetts Lowell \\ {\tt \small \{aromanov,arum\}@cs.uml.edu} \\
$^2$Carnegie Mellon University \\ {\tt \small mdeartea@andrew.cmu.edu,achould@cmu.edu} \\
$^3$Microsoft Research \\ {\tt \small \{wallach,jchayes,Christian.Borgs,Adam.Kalai\}@microsoft.com} \\
$^4$LinkedIn \\ {\tt \small \{sgeyik,kkenthapadi\}@linkedin.com} \\
}

\date{}

\begin{document}
\maketitle
\begin{abstract}
There is a growing body of work that proposes methods for mitigating
bias in machine learning systems. These methods typically rely on
access to protected attributes such as race, gender, or age. However,
this raises two significant challenges: (1) protected attributes may
not be available or it may not be legal to use them, and (2) it is
often desirable to simultaneously consider multiple protected
attributes, as well as their intersections. In the context of
mitigating bias in occupation classification, we propose a method for
discouraging correlation between the predicted probability of an
individual's true occupation and a word embedding of their name. This
method leverages the societal biases that are encoded in word
embeddings, eliminating the need for access to protected
attributes. Crucially, it only requires access to individuals' names
at training time and not at deployment time. We evaluate two
variations of our proposed method using a large-scale dataset of
online biographies. We find that both variations simultaneously reduce
race and gender biases, with almost no reduction in the classifier's
overall true positive rate.
\end{abstract}

\section{Introduction}

\blfootnote{``What's in a name? That which we call a rose by any other name would smell as sweet.'' -- \textit{William Shakespeare, Romeo and Juliet.}}

\blfootnote{Accepted at NAACL 2019.}

In recent years, the performance of machine learning systems has improved substantially, leading to the widespread use of machine learning in many domains, including high-stakes domains such as healthcare, employment, and criminal justice~\cite{chalfin2016productivity, miotto2017deep, Chouldechova17}. This increased prevalence has led many people to ask the question, ``accurate, but for whom?''~\cite{chouldechova2017fairer}.\looseness=-1

When the performance of a machine learning system differs
substantially for different groups of people, a number of concerns
arise~\citep{barocas2016big,kim2016data}. First and foremost, there is a risk that the deployment of such
a method may harm already marginalized groups and widen existing
inequalities. Recent work highlights this concern in the context of
online recruiting and automated hiring~\citep{biasbios}. When
predicting an individual's occupation from their online biography, the
authors show that if occupation-specific gender gaps in true positive
rates are correlated with existing gender imbalances in those
occupations, then those imbalances will be compounded over time---a
phenomenon sometimes referred to as the ``leaky pipeline.'' Second,
the correlations that lead to performance differences between groups
are often irrelevant. For example, while an occupation classifier
should predict a higher probability of software engineer if an
individual's biography mentions coding experience, there is no good
reason for it to predict a lower probability of software engineer if
the biography also mentions softball.




Prompted by such concerns about bias in machine learning systems,
there is a growing body of work on fairness in machine learning. Some
of the foundational papers in this area highlighted the limitations
of trying to mitigate bias using methods that are ``unaware'' of
protected attributes such as race, gender, or
age~\citep[e.g.,][]{Dwork:2012}. As a result, subsequent work has
primarily focused on introducing fairness constraints, defined in
terms of protected attributes, that reduce incentives to rely on
undesirable correlations~\citep[e.g.,][]{hardt2016equality,
  zhang2018mitigating}. This approach is particularly useful if
similar performance can be achieved by slightly different
means---i.e., fairness constraints may
aid in model selection if there are many near-optima.\looseness=-1

In practice, though, any approach that relies on protected attributes
may stand at odds with anti-discrimination law, which limits the use
of protected attributes in domains such as employment and education,
even for the purpose of mitigating bias. And, in other domains,
protected attributes are often not
available~\citep{holstein19}. Moreover, even when they are,
it is usually desirable to simultaneously consider multiple protected
attributes, as well as their intersections. For example,
\citet{buolamwini2017gender} showed that commercial gender
classifiers have higher error rates for women with
darker skin tones than for either women or people with darker skin
tones overall.

We propose a method for reducing bias in machine
learning classifiers without relying on protected attributes. In the
context of occupation classification, this method discourages a
classifier from learning a correlation between the predicted
probability of an individual's occupation and a word embedding of
their name. Intuitively, the probability of an individual's occupation
should not depend on their name---nor on any protected attributes that
may be inferred from it. We present two variations of our
method---i.e., two loss functions that enforce this constraint---and
show that they simultaneously reduce both race and gender biases with
little reduction in classifier accuracy. Although we are motivated by
the need to mitigate bias in online recruiting and automated hiring,
our method can be applied in any domain where individuals' names are
available at training time.\looseness=-1

Instead of relying on protected attributes, our method leverages the
societal biases that are encoded in word
embeddings~\cite{bolukbasi2016man,caliskan2017semantics}. In
particular, we build on the work of~\citet{swinger2018biases}, which
showed that word embeddings of names typically reflect the societal
biases that are associated with those names, including race, gender,
and age biases, as well encoding information about other factors that
influence naming practices such as nationality and religion. By using
word embeddings of names as a tool for mitigating bias, our method is
conceptually simple and empirically powerful. Much like the ``proxy
fairness'' approach of~\citet{gupta2018proxy}, it is applicable when
protected attributes are not available; however, it additionally
eliminates the need to specify which biases are to be mitigated, and
allows simultaneous mitigation of multiple biases, including those
that relate to group intersections. Moreover, our method only requires
access to proxy information (i.e., names) at training time and not at
deployment time, which avoids disparate treatment concerns and extends
fairness gains to individuals with ambiguous names. For example, a
method that explicitly or implicitly infers protected attributes from
names at deployment time may fail to correctly infer that an
individual named Alex is female and, in turn, fail to mitigate gender
bias for her. Methodologically, our work is also similar to that
of~\citet{zafar2017fairness}, which promotes fairness by requiring
that the covariance between a protected attribute and a data point's
distance from a classifier's decision boundary is smaller than some
constant. However, unlike our method, it requires access to protected
attributes, and does not facilitate simultaneous mitigation of
multiple biases.

We present our method in Section~\ref{sec:method}. In
section~\ref{sec:evaluation}, we describe our evaluation, followed by
results in Section~\ref{sec:results} and conclusions in
Section~\ref{sec:conclusion}.

\section{Method}
\label{sec:method}

Our method discourages an
occupation classifier from learning a correlation between the
predicted probability of an individual's occupation and a word
embedding of their name. In this section, we present two variations of
our method---i.e., two penalties that can be added to an arbitrary
loss function and used when training any classifier.

We assume that each data point corresponds to an individual, with a
label indicating that individual's occupation. We also assume access
to the names of the individuals represented in the training set. The
first variation, which we call Cluster Constrained Loss (CluCL), uses
$k$-means to cluster word embeddings of the names in the training
set. Then, for each pair of clusters, it minimizes between-cluster
disparities in the predicted probabilities of the true labels for the
data points that correspond to the names in the clusters. In contrast,
the second variation minimizes the covariance between the predicted
probability of an individual's occupation and a word embedding of
their name. Because this variation minimizes the covariance directly,
we call it Covariance Constrained Loss (CoCL). The most salient
difference between these variations is that CluCL only minimizes
disparities between the latent groups captured by the clusters. For
example, if the clusters correspond only to gender, then CluCL is only
capable of mitigating gender bias. However, given a sufficiently large
number of clusters, CluCL is able to simultaneously mitigate multiple
biases, including those that relate to group intersections. For both
variations, individual's names are not used as input to the classifier
itself; they appear only in the loss function used when training the
classifier. The resulting trained classifier can therefore be deployed
without access to individuals' names.\looseness=-1



\subsection{Formulation}
\label{sec:formualtion}

We define $x_i = \{ x^1_i,\dots, x^M_i\}$ to be a data point, $y_i$ to
be its corresponding (true) label, and $n^f_i$ and $n^l_i$ to be the first
and last name of the corresponding individual. The classification task
is then to (correctly) predict the label for each data point:
\begin{align}
    p_i &= H(x_i) \\ 
    \hat{y}_i &= \argmax_{1 \le j \le |C|} p_i[j],
\end{align}
where $H(\cdot)$ is the classifier, $C$ is the set of possible classes, $p_i \in \mathbb{R}^{|C|}$ is the output of the classifier for data point $x_i$---e.g., $p_i[j]$ is the predicted probability of $x_i$ belonging to class $j$---and $\hat{y}_i$ is the predicted label for $x_i$. We define $p_i^y$ to be the predicted probability of $y_i$---i.e., the true label for $x_i$.


The conventional way to train such a classifier is to minimize some
loss function $\mathcal{L}$, such as the cross-entropy loss
function. Our method simply adds an additional penalty to this loss
function:
\begin{align} 
 \mathcal{L}_\text{total} &= \mathcal{L} + \lambda \cdot \mathcal{L}_\text{CL}, 
\end{align}
where $\mathcal{L}_{\textrm{CL}}$ is either
$\mathcal{L}_\textrm{CluCL}$ or $\mathcal{L}_\textrm{CoCL}$ (defined in Sections~\ref{sec:formualtion_fnc} and~\ref{sec:formualtion_fnd}, respectively), and $\lambda$ is
a hyperparameter that determines the strength of the penalty. This
loss function is only used during training, and plays no role in the
resulting trained classifier. Moreover, it can be used in any standard
setup for training a classifier---e.g., training a deep neural network
using mini-batches and the Adam optimization
algorithm~\cite{kingma2014adam}.

\subsection{Cluster Constrained Loss}
\label{sec:formualtion_fnc}

This variation represents each first name $n_i^f$ and last name
$n_i^l$ as a pair of low-dimensional vectors using a set of pretrained
word embeddings $E$. These are then combined to form a single vector:
\begin{equation}
    n^e_i = \frac{1}{2} \left( E[n^f_i] + E[n^l_i]   \right).
\end{equation}
Using $k$-means~\cite{arthur2007k}, CluCL then clusters the resulting
embeddings into $k$ clusters, yielding a cluster assignment $k_i$ for
each name (and corresponding data point). Next, for each class $c \in
C$, CluCL computes the following average pairwise difference between clusters:
\begin{align}
  l_c &= \frac{1}{k(k-1)}\times{}\notag\\
  &\quad \sum_{u,v=1}^k \left( \frac{1}{N_{c,u}} \sum_{\substack{i:y_i=c,\\k_i = u}} p^y_i - \frac{1}{N_{c,v}} \sum_{\substack{i:y_i=c,\\k_i= v}} p^y_i \right)^2,
\end{align}
where $u$ and $v$ are clusters and $N_{c,u}$ is the
number of data points in cluster $u$ for which $y_i=c$. CluCL considers each class individually because different classes will likely have different numbers of training data points and different disparities. Finally, CluCL computes the average of $l_1, \ldots l_{|C|}$ to yield
\begin{equation}
    \mathcal{L}_\text{CluCL} = \frac{1}{|C|} \sum_{c \in C} l_c.
\end{equation}


\subsection{Covariance Constrained Loss}
\label{sec:formualtion_fnd}

This variation minimizes the covariance between the predicted
probability of a data point's label and the corresponding individual's
name. Like CluCL, CoCL represents each name as a single vector $n_i^e$
and considers each class individually:\looseness=-1
\begin{equation}
    l_c = \mathbb{E}_{i:y_i = c} \left[ \left( p^y_i - \mu_p^c \right) \cdot \left( n^e_i - \mu_n^c \right) \right],
\end{equation}
where $\mu_p^c = \mathbb{E}_{i:y_i=c}\left[ p^y_i \right]$ and $\mu_n^c = \mathbb{E}_{i:y_i = c} \left[ n^e_i \right]$. Finally, CoCL computes the following average:
\begin{equation*}
   \mathcal{L}_\text{CoCL} = \frac{1}{|C|} \sum_{c \in C} \Vert l_c \Vert,
\end{equation*}
where $\Vert \cdot \Vert$ is the $\ell_2$ norm.

\section{Evaluation}
\label{sec:evaluation}

One of our method's strengths is its ability to simultaneously
mitigate multiple biases without access to protected attributes;
however, this strength also poses a challenge for evaluation. We are
unable to quantify this ability without access to these attributes. To
facilitate evaluation, we focus on race and gender biases only because
race and gender attributes are more readily available than attributes
corresponding to other biases. We
further conceptualize both race and gender to be binary
(``white/non-white'' and ``male/female'') but note that these
conceptualizations are unrealistic, reductive simplifications that fail to
capture many aspects of race and gender, and erase anyone who does not
fit within their assumptions. We
emphasize that we use race and gender attributes only for evaluation---they do not play a role in our method.

\subsection{Datasets}

We use two datasets to evaluate our method: the adult income dataset
from the UCI Machine Learning Repository~\cite{Dua:2017}, where the
task is to predict whether an individual earns more than \$50k per
year (i.e., whether their occupation is ``high status''), and a
dataset of online biographies~\cite{biasbios}, where the task is to
predict an individual's occupation from the text of their online
biography.

Each data point in the \textit{Adult} dataset consists
of a set of binary, categorical, and continuous attributes, including
race and gender. We preprocess these attributes to more easily allow
us to understand the classifier's decisions. Specifically, we
normalize continuous attributes to be in the range $[0,1]$ and we
convert categorical attributes into binary indicator
variables. Because the data points do not have names associated with
them, we generate synthetic first names using the race and gender
attributes. First, we use the dataset
of~\citet{tzioumis2018demographic} to identify ``white'' and
``non-white'' names. For each name, if the proportion of ``white''
people with that name is higher than 0.5, we deem the name to be
``white;'' otherwise, we deem it to be ``non-white.''\footnote{For
  90\% of the names, the proportion of ``white'' people with that name
  is greater than 0.7 or less than 0.3, so there is a clear
  distinction between ``white'' and ``non-white'' names.} Next, we use
Social Security Administration data about baby names~\citeyearpar{SSA}
to identify ``male'' and ``female'' names. For each name, if the
proportion of boys with that name is higher than 0.5, we deem the name
to be ``male;'' otherwise, we deem it to be ``female.''\footnote{For
  98\% of the names, the proportion of boys with that name is greater
  than 0.7 or less than 0.3, so there is an even clearer distinction
  between ``male'' and ``female'' names.} We then take the
intersection of these two sets of names to yield a single set of names
that is partitioned into four non-overlapping categories by (binary)
race and gender. Finally, we generate a synthetic first name for each
data point by sampling a name from the relevant category.

Each data point in the \textit{Bios} dataset consists of the text of
an individual's biography, written in the third person. We represent
each biography as a vector of length $V$, where $V$ is the size of the
vocabulary. Each element corresponds to a single word type and is
equal to 1 if the biography contains that type (and 0 otherwise). We
limit the size of the vocabulary by discarding the 10\% most common
word types, as well as any word types that occur fewer than twenty
times. Unlike the \textit{Adult} dataset, each data point has a name
associated with it. And, because biographies are typically written in
the third person and because pronouns are gendered in English, we can
extract (likely) self-identified gender. We infer race for each data
point by sampling from a Bernoulli distribution with probability equal
to the average of the probability that an individual with that first
name is ``white'' (from the dataset
of~\citet{tzioumis2018demographic}, using a threshold of 0.5, as
described above) and the probability that an individual with that last
name is ``white'' (from the dataset of~\citet{comenetz2016frequently},
also using a threshold of 0.5).\footnote{We note that, in general, an individual's race
  or gender should be directly reported by the individual in question;
  inferring race or gender can be both inaccurate and reductive.} Finally, like~\citet{biasbios}, we
consider two versions of the \textit{Bios} dataset: one where first
names and pronouns are available to the classifier and one where they
are ``scrubbed.''

Throughout our evaluation, we use the fastText word embeddings,
pretrained on Common Crawl data~\cite{bojanowski2016enriching}, to
represent names.

\subsection{Classifier and Loss Function}
\label{sec:classifier}

Our method can be used with any classifier, including deep neural
networks such as recurrent neural networks and convolutional neural
networks. However, because the focus of this paper is mitigating bias,
not maximizing classifier accuracy, we use a single-layer neural
network:\looseness=-1
\begin{align*}
h_i &= W_h \cdot x_i + b_h \\
p_i &= \text{softmax}(h_i)
\end{align*}
where $W_h \in \mathbb{R}^{|C| \times M}$ and $b_h \in
\mathbb{R}^{|C|}$ are the weights. This structure allows us to
examine individual elements of the matrix $W_h$ in order to understand
the classifier's decisions for any dataset.

Both the \textit{Adult} dataset and the \textit{Bios} dataset have a
strong class imbalance. We therefore use a weighted cross-entropy loss as
$\mathcal{L}$, with weights set to the values proposed by~\citet{king01}.

\subsection{Quantifying Bias}

To quantify race bias and gender bias, we follow the approach proposed
by~\citet{biasbios} and compute the true positive rate (TPR) race gap
and the TPR gender gap---i.e., the differences in the TPRs between
races and between genders, respectively---for each occupation. The TPR
race gap for occupation $c$ is defined as follows:
\begin{align}
\text{TPR}_{r,c} &= P\left[ \hat{Y} = c \,|\, R=r, Y=c \right] \\   
\text{Gap}_{r,c} &= \text{TPR}_{r,c} - \text{TPR}_{\sim r,c},
\end{align}
where $r$ and ${\sim} r$ are binary races, $\hat{Y}$ and $Y$ are
random variables representing the predicted and true occupations for
an individual, and $R$ is a random variable representing that
individual's race. Similarly, the TPR gender gap for occupation $c$ is\looseness=-1
\begin{align}
\text{TPR}_{g,c} &= P\left[ \hat{Y} = c \,|\, G=g, Y=c \right] \\   
\text{Gap}_{g,c} &= \text{TPR}_{g,c} - \text{TPR}_{\sim g,c},
\end{align}
where $g$ and ${\sim} g$ are binary genders and $G$ is a random
variable representing an individual's gender.

To obtain a single score that quantifies race bias, thus facilitating
comparisons, we calculate the root mean square of the per-occupation
TPR race gaps:
\begin{equation}
    \text{Gap}^{\text{RMS}}_r = \sqrt{ \frac{1}{|C|} \sum_{c \in C} \text{Gap}_{r,c}^2 }.
\end{equation}
We obtain a single score that quantifies gender bias similarly. The
motivation for using the root mean square instead of an average is
that larger values have a larger effect and we are more interested in
mitigating larger biases. Finally, to facilitate worst-case
analyses, we calculate the maximum TPR race gap and the maximum
TPR gender gap.

We again emphasize that race and gender attributes are used only for
evaluating our method.



    

\begin{figure*}[ht]
\centering
\begin{subfigure}{.4\textwidth}
    \centering
    \includegraphics[width=\linewidth]{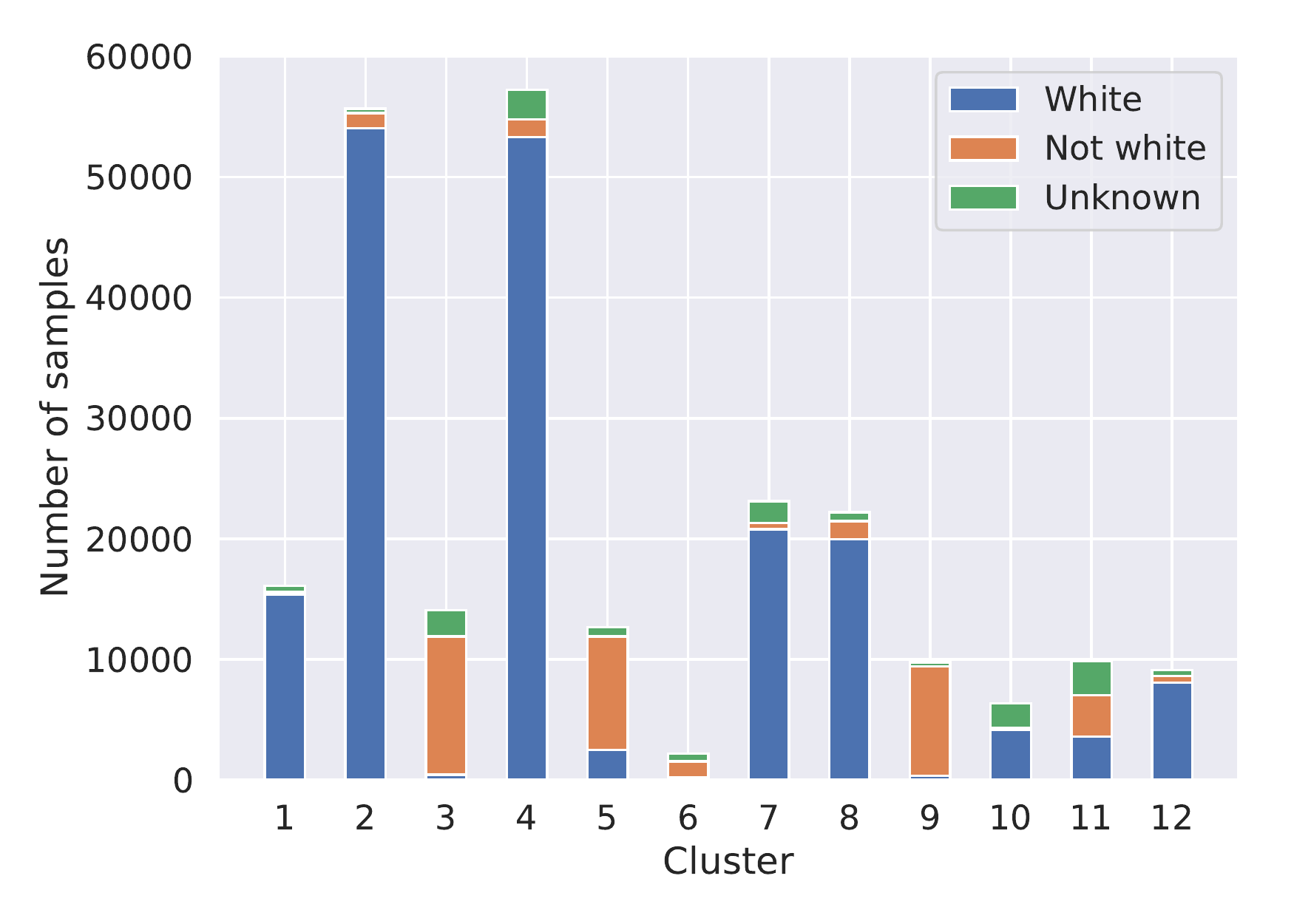}
    \caption{Race.}
    \label{fig:bios_clusters_races}
\end{subfigure}%
\begin{subfigure}{.4\textwidth}
  \centering
    \includegraphics[width=\linewidth]{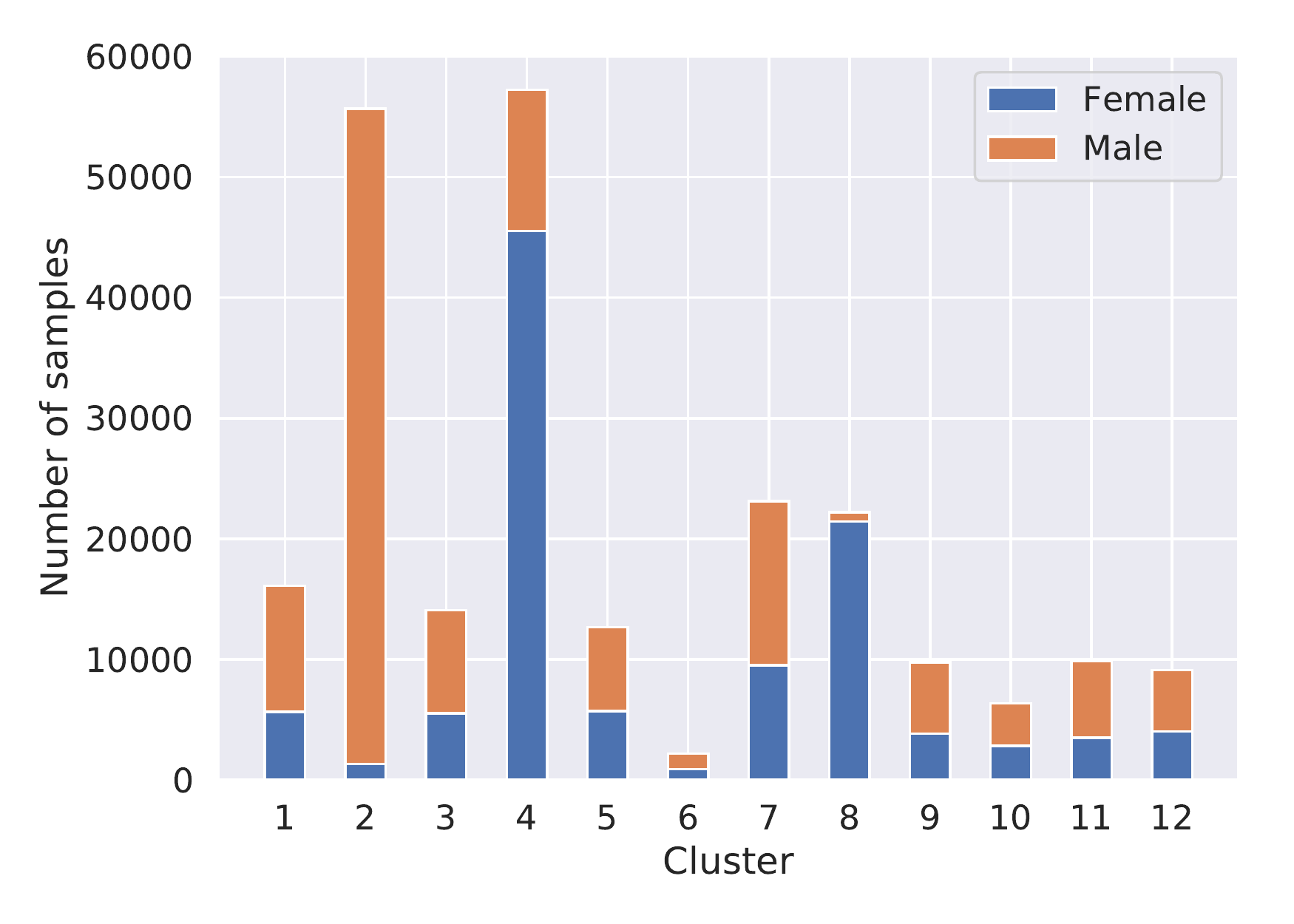}
    \caption{Gender.}
    \label{fig:bios_clusters_genders}
\end{subfigure}
\caption{Number of data points (from the \textit{Bios} dataset) in
  each cluster that correspond to each race and gender.}
\label{fig:bios_clusters}
\end{figure*}

\section{Results}
\label{sec:results}

We first demonstrate that word embeddings of names encode information
about race and gender. We then present the main results of our evaluation,
before examining individual elements of the matrix $W_h$ in order to
better understand our method.

\subsection{Word Embeddings of Names as Proxies}

We cluster the names associated with the data points in the
\textit{Bios} dataset, represented as word embeddings, to verify that
such embeddings indeed capture information about race and
gender. We perform $k$-means clustering (using the $k$-means++
algorithm) with $k=12$ clusters, and then plot the number of data
points in each cluster that correspond to each (inferred) race and
gender. Figures~\ref{fig:bios_clusters_races}
and~\ref{fig:bios_clusters_genders} depict these numbers,
respectively.\looseness=-1

Clusters 1, 2, 4, 7, 8, and 12 contain mostly ``white'' names, while
clusters 3, 5, and 9 contain mostly ``non-white names.'' Similarly,
clusters 4 and 8 contain mostly ``female'' names, while cluster 2
contains mostly ``male'' names. The other clusters are more balanced
by race and gender. Manual inspection of the clusters reveals that
cluster 9 contains mostly Asian names, while cluster 8 indeed contains
mostly ``female'' names. The names in cluster 2 are mostly ``white''
and ``male,'' while the names in cluster 4 are mostly ``white'' and
``female.'' This suggests that the clusters are capturing at least
some intersections. Together these results demonstrate that word
embeddings of names do indeed encode at least some information about
race and gender, even when first and last names are combined into a
single embedding vector. For a longer discussion of the societal
biases reflected in word embeddings of names, we recommend the work
of~\citet{swinger2018biases}.

\subsection{\textit{Adult} Dataset}

\begin{table*}[ht]
\centering
\small
\begin{tabular}{@{}rcccccc@{}}
\toprule
Method & \multicolumn{1}{c}{$\lambda$} & \multicolumn{1}{c}{Balanced TPR} & \multicolumn{1}{c}{$\textrm{Gap}_g^{\text{RMS}}$} & \multicolumn{1}{c}{$\textrm{Gap}_r^{\text{RMS}}$} & \multicolumn{1}{c}{$\textrm{Gap}_g^{\text{max}}$} & \multicolumn{1}{c}{$\textrm{Gap}_r^{\textrm{max}}$} \\ \midrule
None & 0 & \textbf{0.795} & 0.299 & 0.120 & 0.303 & 0.148 \\ \midrule
CluCL & 1 & 0.788 & 0.278 & 0.121 & 0.297 & 0.145 \\
CluCL & 2 & 0.793 & 0.259 & 0.085 & 0.282 & 0.114 \\ \midrule
CoCL & 1 & 0.794 & 0.215 & 0.091 & 0.251 & 0.119 \\
CoCL & 2 & 0.790 & \textbf{0.163} & \textbf{0.080} & \textbf{0.201} & \textbf{0.109} \\ \bottomrule
\end{tabular}
\caption{Results for the \textit{Adult} dataset. Balanced TPR (i.e., per-occupation TPR, averaged over occupations), gender bias quantified as $\textrm{Gap}_g^{\text{RMS}}$, race bias quantified as $\textrm{Gap}_r^{\text{RMS}}$, maximum TPR gender gap, and maximum TPR race gap for different values of hyperparameter $\lambda$. Results are averaged over four runs with different random seeds.}
\label{tbl:results_adult}
\end{table*}

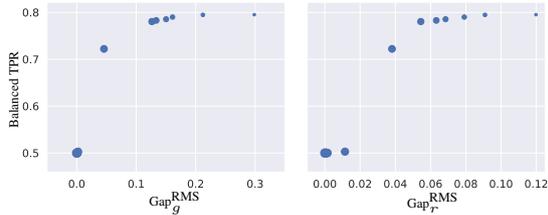
\begin{figure}[t]

\begin{tikzpicture}
    \node[anchor=south west,inner sep=0] (image) at (0,0) {
        \includegraphics[width=\linewidth]{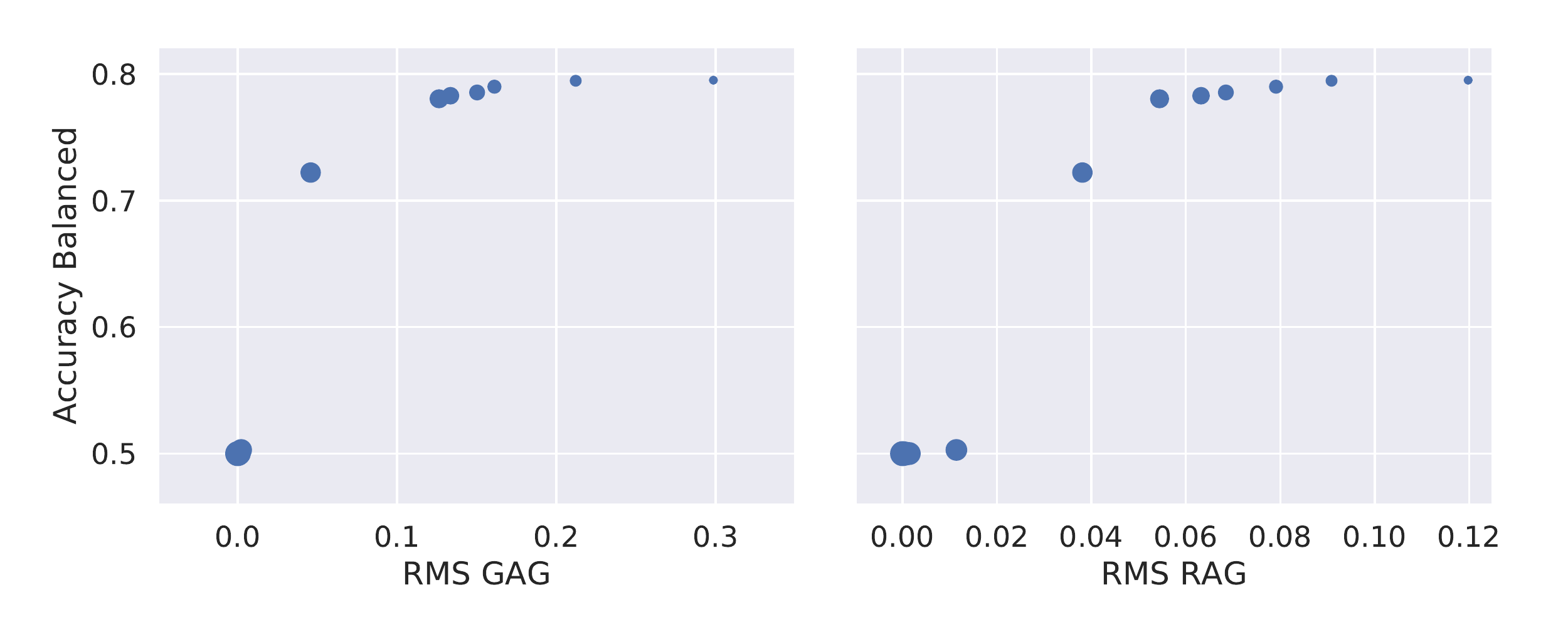}
    };
    \begin{scope}[x={(image.south east)},y={(image.north west)}]
        \fill [white] (0.1,0.05) rectangle (0.97,0.12);
        \node  at (0.32,0.065) {{\fontsize{5}{6}\selectfont $\textrm{Gap}_g^{\text{RMS}}$}};
        \node  at (0.76,0.065) {{\fontsize{5}{6}\selectfont $\textrm{Gap}_r^{\text{RMS}}$}};
        
        \fill [white] (0.01,0.1) rectangle (0.055,0.95);
        \node [rotate=90] at (0.045,0.55) {{\fontsize{5}{6}\selectfont Balanced TPR}};
    \end{scope}
\end{tikzpicture}

\caption{Gender bias quantified as $\textrm{Gap}_g^{\text{RMS}}$ (left) and race bias quantified as $\textrm{Gap}_r^{\textrm{RMS}}$ (right) versus balanced TPR for the CoCL variation of our method with different values of hyperparameter $\lambda$ (a larger dot means a larger value of $\lambda$) for the \textit{Adult} dataset. Results are averaged over four runs with different random seeds.}
\label{fig:adult_gaps_rms}
\end{figure}

\begin{table*}[ht]
\centering
\small
\begin{tabular}{@{}rcccccc@{}}
\toprule
Method & \multicolumn{1}{c}{$\lambda$} & \multicolumn{1}{c}{Balanced TPR} & \multicolumn{1}{c}{$\textrm{Gap}_g^{\text{RMS}}$} & \multicolumn{1}{c}{$\textrm{Gap}_r^{\text{RMS}}$} & \multicolumn{1}{c}{$\textrm{Gap}_g^{\text{max}}$} & \multicolumn{1}{c}{$\textrm{Gap}_r^{\textrm{max}}$} \\ \midrule
None & 0 & \textbf{0.788} & 0.173 & 0.051 & 0.511 & 0.121 \\ \midrule
CluCL & 1 & 0.784 & 0.168 & 0.048 & 0.494 & 0.120 \\
CluCL & 2 & 0.781 & \textbf{0.165} & \textbf{0.047} & \textbf{0.486} & 0.114 \\ \midrule
CoCL & 1 & 0.785 & 0.168 & 0.048 & 0.507 & \textbf{0.109} \\
CoCL & 2 & 0.779 & 0.169 & 0.048 & 0.512 & 0.116 \\ \bottomrule
\end{tabular}
\caption{Results for the original \textit{Bios} dataset. Balanced TPR (i.e., per-occupation TPR, averaged over occupations), gender bias quantified as $\textrm{Gap}_g^{\text{RMS}}$, race bias quantified as $\textrm{Gap}_r^{\text{RMS}}$, maximum TPR gender gap, and maximum TPR race gap for different values of hyperparameter $\lambda$. Results are averaged over four runs with different random seeds.}
\label{tbl:results_bios_no_scrubbed}
\end{table*}

\begin{table*}[ht]
\centering
\small
\begin{tabular}{@{}rcccccc@{}}
\toprule
Method & \multicolumn{1}{c}{$\lambda$} & \multicolumn{1}{c}{Balanced TPR} & \multicolumn{1}{c}{$\textrm{Gap}_g^{\text{RMS}}$} & \multicolumn{1}{c}{$\textrm{Gap}_r^{\text{RMS}}$} & \multicolumn{1}{c}{$\textrm{Gap}_g^{\text{max}}$} & \multicolumn{1}{c}{$\textrm{Gap}_r^{\textrm{max}}$} \\ \midrule
None & 0 & \textbf{0.785} & 0.111 & 0.049 & 0.385 & 0.123 \\ \midrule
CluCL & 1 & 0.782 & \textbf{0.107} & 0.048 & \textbf{0.383} & 0.112 \\
CluCL & 2 & 0.778 & 0.112 & 0.046 & 0.395 & \textbf{0.107} \\ \midrule
CoCL & 1 & 0.780 & 0.109 & 0.047 & 0.388 & 0.117 \\
CoCL & 2 & 0.775 & 0.108 & \textbf{0.046} & 0.387 & 0.109 \\ \bottomrule
\end{tabular}
\caption{Results for the ``scrubbed'' \textit{Bios} dataset. Balanced TPR (i.e., per-occupation TPR, averaged over occupations), gender bias quantified as $\textrm{Gap}_g^{\text{RMS}}$, race bias quantified as $\textrm{Gap}_r^{\text{RMS}}$, maximum TPR gender gap, and maximum TPR race gap for different values of hyperparameter $\lambda$. Again, results are averaged over four runs.}
\label{tbl:results_bios_scrubbed}
\end{table*}

The results of our evaluation using the \textit{Adult} dataset are shown in 
Table~\ref{tbl:results_adult}. The task is to predict whether an
individual earns more than \$50k per year (i.e., whether their
occupation is ``high status''). Because the dataset has a strong class
imbalance, we report the balanced TPR---i.e., we compute the per-class
TPR and then average over the classes. We experiment with different
values of the hyperparameter $\lambda$. When $\lambda=0$, the method is equivalent
to using the conventional weighted cross-entropy loss function. Larger
values of $\lambda$ increase the strength of the penalty, but may lead to a
less accurate classifier. Using $\lambda=0$ leads to significant gender
bias: the maximum TPR gender gap is 0.303. This means that the TPR
is 30\% higher for men than for women. We emphasize that this does
\emph{not} mean that the classifier is more likely to predict that a
man earns more than \$50k per year, but means that the classifier is
more likely to \emph{correctly} predict that a man earns more than
\$50k per year. Both variations of our method significantly reduce race and
gender biases. With CluCL, the root mean square TPR race gap is
reduced from 0.12 to 0.085, while the root mean square TPR gender gap
is reduced from 0.299 to 0.25. These reductions in bias result in less
than one percent decrease in the balanced TPR (79.5\% is decreased to
79.3\%). With CoCL, the race and gender biases are further
reduced: the root mean square TPR race gap is reduced to 0.08, while
the root mean square TPR gender gap is reduced to 0.163, with 0.5\%
decrease in the balanced TPR.

We emphasize that although our proposed method significantly reduces
race and gender biases, neither variation can completely eliminate
them. In order to understand how different values of hyperparameter
$\lambda$ influence the reduction in race and gender biases, we perform
additional experiments using CoCL where we vary $\lambda$ from 0 to
10. Figure~\ref{fig:adult_gaps_rms} depicts these results. Larger
values of $\lambda$ indeed reduce race and gender biases; however, to
achieve a root mean square TPR gender gap of zero means reducing the
balanced TPR to 50\%, which is unacceptably low. That said, there are
a wide range of values of $\lambda$ that significantly reduce race and
gender biases, while maintaining an acceptable balanced TPR. For
example, $\lambda=6$ results in a root mean square TPR race gap of 0.038 and
a root mean square TPR gender gap of 0.046, with only a 7.3\% decrease
in the balanced TPR.

\subsection{\textit{Bios} Dataset}

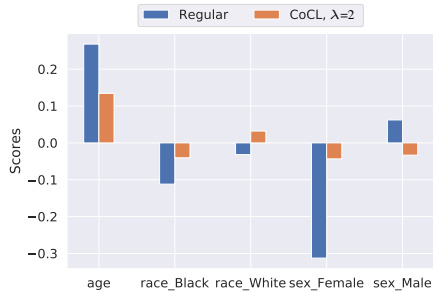
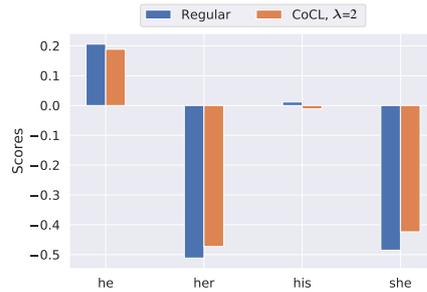
\begin{figure*}[ht]
\centering
\begin{subfigure}{.4\textwidth}
    \centering

    \begin{tikzpicture}
        \definecolor{imgbackcolor}{RGB}{238,238,244}
        \node[anchor=south west,inner sep=0] (image) at (0,0) {
            \includegraphics[width=\linewidth]{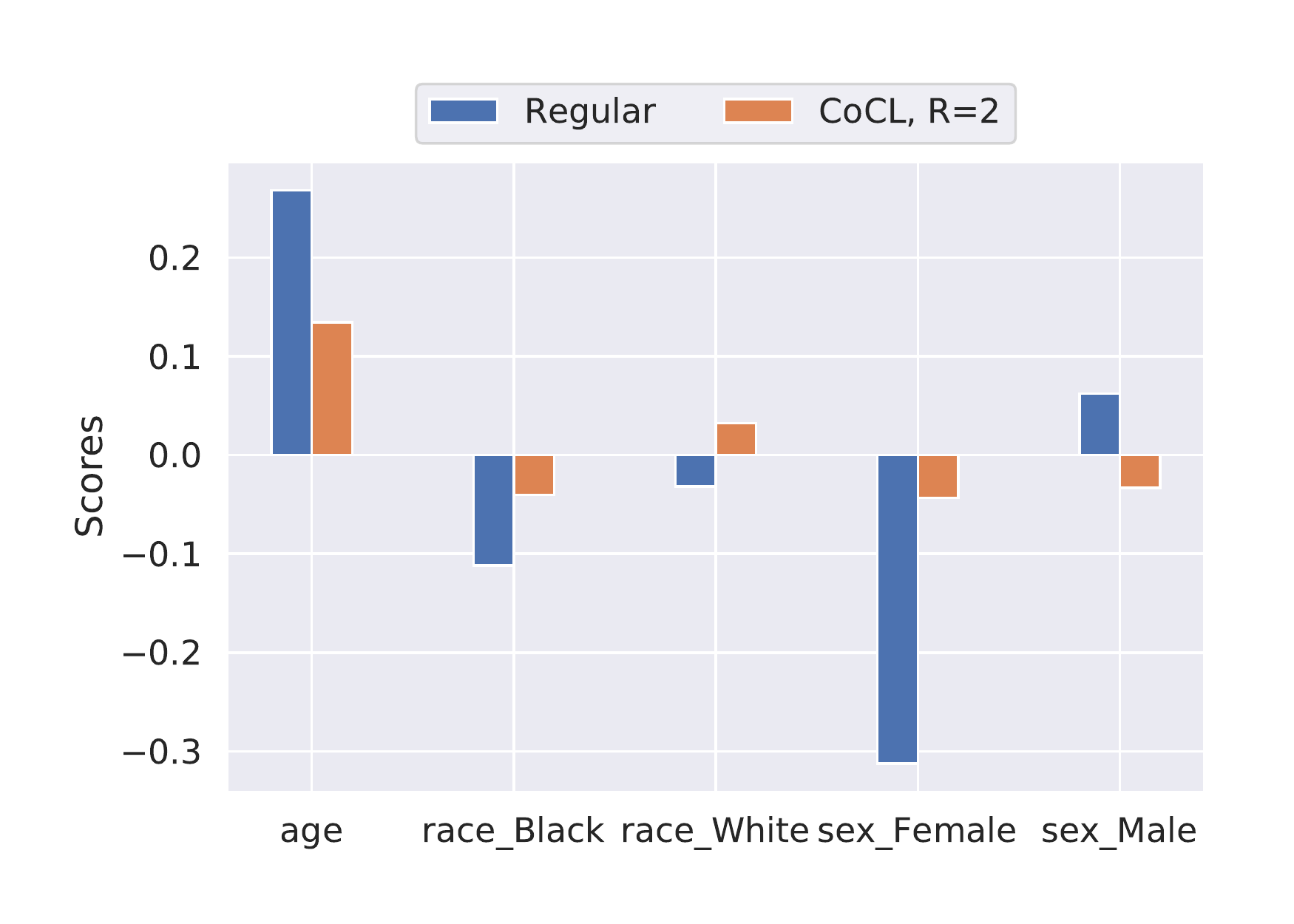}
        };
        \begin{scope}[x={(image.south east)},y={(image.north west)}]
            \fill [imgbackcolor] (0.715,0.85) rectangle (0.78,0.90);
            \node  at (0.74,0.88) {{\fontsize{5}{5}\selectfont $\lambda$=2}};
        \end{scope}
    \end{tikzpicture}

    \caption{\textit{Adult} dataset.}
    \label{fig:results_wegith_analysis_adult}
\end{subfigure}%
\begin{subfigure}{.4\textwidth}
    \centering

    \begin{tikzpicture}
        \definecolor{imgbackcolor}{RGB}{238,238,244}
        \node[anchor=south west,inner sep=0] (image) at (0,0) {
            \includegraphics[width=\linewidth]{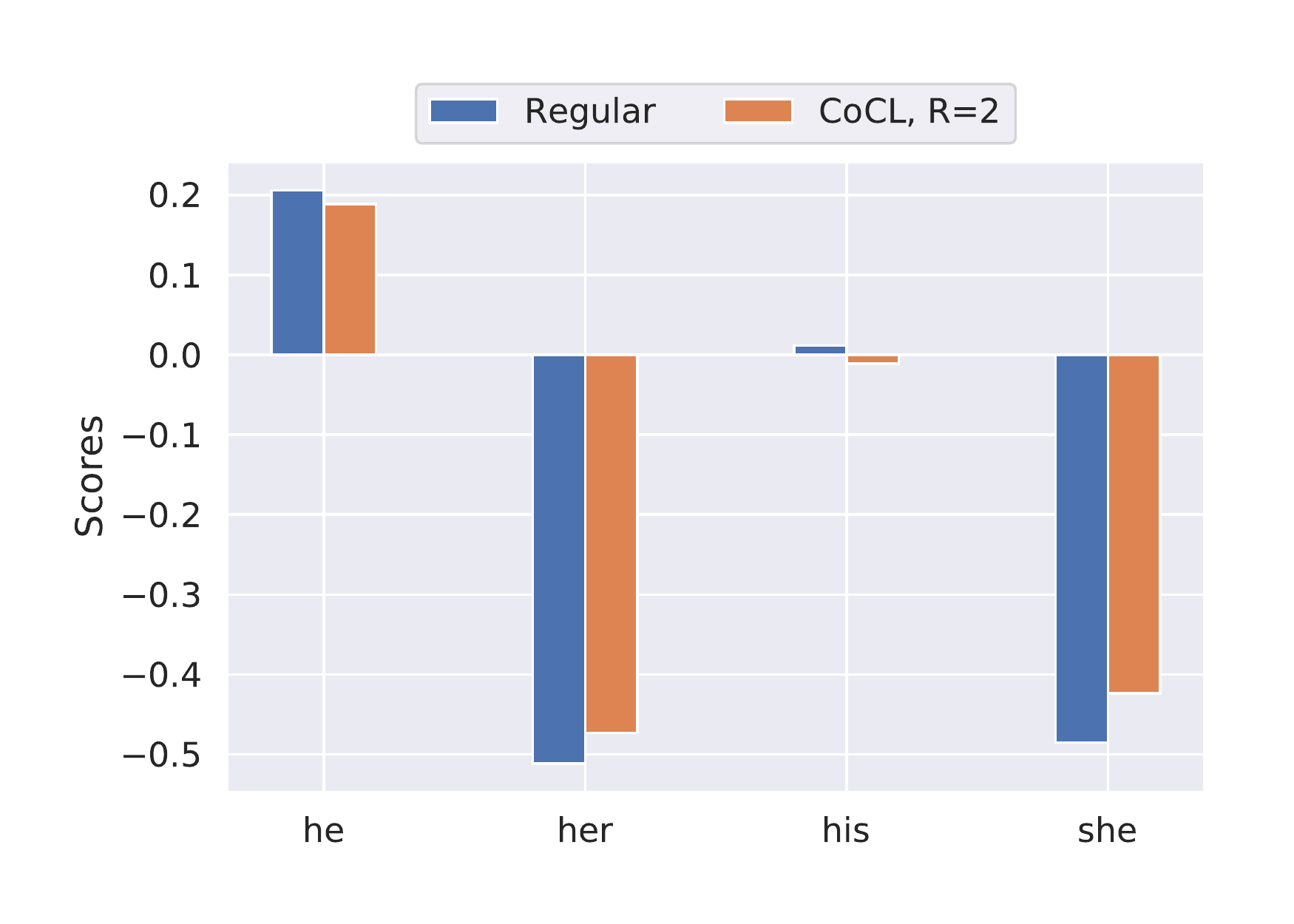}
        };
        \begin{scope}[x={(image.south east)},y={(image.north west)}]
            \fill [imgbackcolor] (0.715,0.85) rectangle (0.78,0.90);
            \node  at (0.74,0.88) {{\fontsize{5}{5}\selectfont $\lambda$=2}};
        \end{scope}
    \end{tikzpicture}

    \caption{\textit{Bios} dataset, occupation ``surgeon.''}
    \label{fig:results_wegith_analysis_bios}
\end{subfigure}
\caption{Classifier weight values for several attributes for the conventional weighted cross-entropy loss function (i.e., $\lambda=0$) and for CoCL with $\lambda=2$. Results are averaged over four runs with different random seeds.}
\label{fig:results_wegith_analysis}
\end{figure*}

The results of our evaluation using the original and ``scrubbed''
(i.e., names and pronouns are ``scrubbed'') versions of the
\textit{Bios} dataset are shown in
Tables~\ref{tbl:results_bios_no_scrubbed}
and~\ref{tbl:results_bios_scrubbed}, respectively. The task is to
predict an individual's occupation from the text of their online
biography. Because the dataset has a strong class imbalance, we again
report the balanced TPR. CluCL and CoCL reduce race and gender biases
for both versions of the dataset. For the original version, CluCL
reduces the root mean square TPR gender gap from 0.173 to 0.165 and
the maximum TPR gender gap by 2.5\%. Race bias is also reduced, though
to a lesser extent. These reductions reduce the balanced TPR by
0.7\%. For the ``scrubbed'' version, the reductions in race and gender
biases are even smaller, likely because most of the information about
race and gender has been removed by ``scrubbing'' names and
pronouns. We hypothesize that these smaller reductions in race and
gender biases, compared to the \textit{Adult} dataset, are because the
\textit{Adult} dataset has fewer attributes and classes than the
\textit{Bios} dataset, and contains explicit race and gender
information, making the task of reducing biases much simpler. We also
note that each biography in the \textit{Bios} dataset is represented
as a vector of length $V$, where $V$ is over 11,000. This means that
the corresponding classifier has a very large number of weights,
and there is a strong overfitting effect. Because this overfitting
effect increases with $\lambda$, we suspect it explains why CluCL has a
larger root mean square TPR gender gap when $\lambda=2$ than when
$\lambda=1$. Indeed, the root mean square TPR gender gap for the training
set is 0.05 when $\lambda=2$. Using dropout and $\ell_2$ weight
regularization lessened this effect, but did not eliminate it entirely.

\subsection{Understanding the Method}
\label{sec:results_wegith_analysis}

Our method mitigates bias by making training-time adjustments to the
classifier's weights that minimize the correlation between the
predicted probability of an individual's occupation and a word
embedding of their name. Because of our choice of classifier (a
single-layer neural network, as described in
Section~\ref{sec:classifier}), we can examine individual elements of
the matrix $W_h$ to understand the effect of our method on the
classifier's decisions. Figure~\ref{fig:results_wegith_analysis_adult}
depicts the values of several weights for the conventional weighted
cross-entropy loss function (i.e., $\lambda=0$) and for CoCL with $\lambda=2$ for
the \textit{Adult} dataset. When $\lambda=0$, the attributes ``sex\_Female''
and ``sex\_Male'' have large negative and positive weights,
respectively. This means that the classifier is more likely to predict
that a man earns more than \$50k per year. With CoCL, these weights
are much closer to zero. Similarly, the weights for the race
attributes are also closer to zero. We note that the weight for the
attribute ``age'' is also reduced, suggesting that CoCL may have
mitigated some form of age bias.

Figure~\ref{fig:results_wegith_analysis_bios} depicts the values of
several weights specific to the occupation ``surgeon'' for the
conventional weighted cross-entropy loss function (i.e., $\lambda=0$) and
for CoCL with $\lambda=2$ for the original version of the \textit{Bios}
dataset. When $\lambda=0$, the attributes ``she'' and ``her'' have large
negative weights, while the attribute ``he'' has a positive
weight. This means that the classifier is less likely to predict that
a biography that contains the words ``she'' or ``her'' belongs to a
surgeon. With CoCL, these magnitudes of these weights are reduced,
though these reductions are not as significant as the reductions shown
for the \textit{Adult} dataset.


\section{Conclusion}
\label{sec:conclusion}

In this paper, we propose a method for reducing bias in machine
learning classifiers without relying on protected attributes. In
contrast to previous work, our method eliminates the need to specify
which biases are to be mitigated, and allows simultaneous mitigation of
multiple biases, including those that relate to group
intersections. Our method leverages the societal biases that are
encoded in word embeddings of names. Specifically, it discourages an
occupation classifier from learning a correlation between the
predicted probability of an individual's occupation and a word
embedding of their name. We present two variations of our method, and
evaluate them using a large-scale dataset of online biographies. We
find that both variations simultaneously reduce race and gender
biases, with almost no reduction in the classifier's overall true
positive rate. Our method is conceptually simple and empirically
powerful, and can be used with any classifier, including deep neural
networks. Finally, although we focus on English, we expect our method
will work well for other languages, but leave this direction for
future work.

\bibliographystyle{acl_natbib}
\bibliography{main}

\appendix



\end{document}